\documentclass[12pt, final]{l4dc2023}

\usepackage{bm}
\hypersetup{colorlinks,linkcolor=blue,urlcolor=blue}
\usepackage{cleveref}
\newcommand{\R}{\mathbb{R}}

\title[Agile Catching with Whole-Body MPC and Blackbox Policy Learning]{Agile Catching with Whole-Body MPC and Blackbox Policy Learning}
\usepackage{times}

 \author{\Name{Saminda Abeyruwan, 
  Alex Bewley, 
  Nicholas M. Boffi,
  Krzysztof Choromanski,
  David D'Ambrosio,
  Deepali Jain,
  Pannag Sanketi,
 Anish Shankar, 
 Vikas Sindhwani, 
Sumeet Singh,
Jean-Jacques Slotine, 
and Stephen Tu$^*$} \\
\Email{Corresponding author: ssumeet@google.com}  \\
   \addr Robotics at Google, $^*$Alphabetical order}

\begin{document}

\maketitle

\begin{abstract}
    We address a benchmark task in agile robotics: catching objects thrown at high-speed. This is a challenging task that involves tracking, intercepting, and cradling a thrown object with access only to visual observations of the object and the proprioceptive state of the robot, all within a fraction of a second. We present the relative merits of two fundamentally different solution strategies: (i) Model Predictive Control using accelerated constrained trajectory optimization, and (ii) Reinforcement Learning using zeroth-order optimization. We provide insights into various performance trade-offs including sample efficiency, sim-to-real transfer, robustness to distribution shifts, and whole-body multimodality via extensive on-hardware experiments. We conclude with proposals on fusing ``classical'' and ``learning-based'' techniques for agile robot control. Videos of our experiments may be found here: \url{https://sites.google.com/view/agile-catching}.
\end{abstract}

\begin{figure}[h]
  \begin{center}
    \includegraphics[width=1.0\textwidth, height=3.5cm]{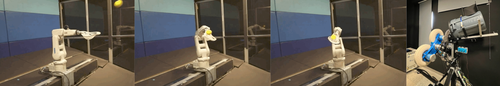}
  \end{center}
  \caption{Mobile Manipulator with Lacrosse Head catching a ball within a second. (right) Automatic ball thrower with controllable yaw angles and speed of around 5m/s.}
  \end{figure}

\section{Introduction}
\label{sec:intro}

Chasing a ball in flight and completing a dramatic diving  catch is a memorable moment of athleticism - a benchmark of human agility - in several popular sports. In this paper, we consider the task of tracking, intercepting and catching balls moving at high speeds on a mobile manipulator platform (see Figure 1), whose end-effector is equipped with a Lacrosse head. Within a fraction of a second, the robot must start continuously translating visual observations of the ball into feasible whole body motions, controlling both the base and the arm in a coordinated fashion. In the final milliseconds, the control system must be robust to perceptual occlusions while also executing a cradling maneuver to stabilize the catch and prevent bounce-out. The physics of this task can be surprisingly complex: despite its geometric simplicity, a ball in flight can swing and curve in unpredictable ways due to drag and Magnus effects~\citep{mehta1985aerodynamics}; furthermore, the contact interaction between the ball and the deformable end-effector involves complex soft-body physics which is challenging to model accurately. 

In this paper, we study the relative merits of synthesizing high speed visual feedback controllers for this task from two ends of a design spectrum: Model Predictive Control (MPC)~\citep{borrelli2017predictive,rawlings2000tutorial} representing a ``pure control" strategy, and Blackbox policy optimization~\citep{choromanski2018structured} representing a ``pure learning" approach. MPC optimizes robot trajectories in real time in response to state uncertainty - it is nearly ``zero-shot" in terms of data requirements and gracefully handles kinematics, dynamics and task-specific constraints, but can be computationally expensive and sensitive to errors in dynamics modeling. On the other hand, policy learning via blackbox or RL (Reinforcement Learning) methods can be extremely data inefficient, but can adapt, in principle, to complex and unknown real world dynamics.  Our primary contribution is to provide insights into subtle trade-offs in reaction time, sample efficiency, robustness to distribution shift, and versatility in terms of whole-body multimodal behaviors in a unified experimental evaluation of robot agility. We conclude the paper with proposals to combine the ``best of both worlds" in future work.

\subsection{Related work}
Both classes of techniques have been previously applied to the robotic catching task. Examples of optimization-based control for ball catching include \cite{Hove1991ExperimentsIR,Hong1995ExperimentsIH, Yu2021NeuralMP,Frese2001OfftheshelfVF,Kober2012PlayingCA}. \cite{Buml2010KinematicallyOC} and \cite{Lampariello2011TrajectoryPF} present an unified approach subsuming catch point selection, catch configuration computation and path generation in a single, nonlinear optimization problem (also see, \cite{Ko2018OnlineOT}, \cite{Jia2019BattingAI}).  Several papers utilize human demonstration and machine learning for parts of the control stack. \cite{Kim2014CatchingOI} probabilistically predict various feasible catching configurations and develop controllers to guide hand–arm motion, which is learned from human demonstration. \cite{Riley2002RobotCT} also learn motion primitives from human demonstration and generate new movements. 
\cite{Dong2020CatchTB} use bi-level motion planning plus a learning based tracking controller.
Some papers aim for soft catching explicitly. \cite{Salehian2016ADS} extend \cite{Kim2014CatchingOI} further, offering a soft catching procedure that is more resilient to imprecisions in controlling the arm and desired time of catch. \cite{Buml2011CatchingFB} extend \cite{Buml2010KinematicallyOC} further for enabling soft landing. \cite{Hong1995ExperimentsIH,Lippiello20123DMR} add heuristics for soft catching, moving the hand along the predicted path of the ball, while decreasing its velocity to allow the dissipation of the impact energy.

\section{Problem formulation and proposed solution}
\label{sec:problem}

We describe the trajectory of the object to be caught  by a function $F_o$, which maps a query time $t \in \R_{\geq 0}$ to the object's position and velocity at time $t$, i.e., $(p_o(t), v_o(t)) \in \R^{3} \times \R^3$. Depending on the aerodynamic and inertial properties of the object, $F_o$ may be highly non-trivial. Our knowledge of $F_o$ is encoded via a known 
 $\hat{F}_o$ which maps a query time $t \in \R_{\geq 0}$ and a set of parameters $\theta_o \in \R^d$ to a prediction for the object position and velocity at time $t$, i.e., $(\hat{p}_o(t; \theta_o), \hat{v}_o(t; \theta_o))$.
 For this work, we limit our scope to spherical, rigid balls and implement $\hat{F}_o$ via classical Newtonian physics; catching objects with non-trivial aerodynamics and non-uniform shapes is left to future work.
However, we only observe the ball position and velocity indirectly via two fixed cameras, and use $\theta_o$ to encode our vision system's current position and velocity estimate.

For the robot, we let $q \in \R^7$ denote the joint configuration vector, where $q_1 \in \R$ corresponds to the translational base joint, and $q_{2:7} \in \R^6$ represent the arm joint angles. 
We also let $\mathtt{FK}: q \in \R^7 \mapsto \mathtt{FK}(q) = (p_h(q), R_h(q)) \in \R^3 \times SO(3)$ denote the forward-kinematics transform that maps the joint configuration vector $q$ to the $SE(3)$ pose of the robot's end-effector, i.e., a lacrosse head.

\subsection{Catching via optimal control}

We assume that there exists a lower-level position and/or velocity controller that compensates for the arm's nonlinear manipulator dynamics. Abstracting away the closed-loop behavior of this lower-level control system, we plan for the motion of the arm by assuming second-order integrator dynamics\footnote{Note that the lower-level control system may have some non-trivial closed-loop response characteristics, including delays. However, these can be pre-compensated for by adjusting the \emph{commanded} $(q, \dot{q})$ trajectory from the \emph{planned} $(q, \dot{q})$ trajectory.} for $q$, i.e., $\ddot{q}(t) = u_a(t) \in \R^7$.

With this assumption, the optimal catching problem (OCP) can be formalized as a \emph{free-end-time constrained optimal control problem} over the function $u_a(\cdot)$ and catching time $t_f$:
\begin{equation}
    \operatorname*{minimize}_{u_a(\cdot), t_f} \ \   J(u_a, t_f) := \int_{0}^{t_f} \left( \lambda + \|u_a(\tau)\|^2 \right)\ d\tau + \Psi(q(t_f), \dot{q}(t_f), t_f),
\label{ocp_obj}
\end{equation}
where $\lambda \in \R_{>0}$ is a weighting constant and $\Psi: \R^7 \times \R^7 \times \R_{\geq 0} \rightarrow 0$ is a terminal cost; subject to the second-order integrator dynamics $\ddot{q}(t) = u_a(t)$, and the following constraints:
\begin{align}
    \forall \tau \in [0, t_f], \ \ u_a(\tau) \in [\underline{u}_a, \overline{u}_a],\ \ q(\tau) &\in [\underline{q}, \overline{q}], \ \ \dot{q}(\tau) \in [\underline{\dot{q}}, \overline{\dot{q}}], \quad \ \  
    c(q(t_f), t_f) \geq 0.
\end{align}
The first three constraints capture limits on the control effort and the joint configurations and velocities. The terminal cost $\Psi$ and endpoint constraint function $c$ capture two desirable properties for catching:
(i) $SE(3)$ pose alignment of the lacrosse head with the ball's position and velocity direction at the catching time $t_f$, and (ii) minimizing any residual velocity of the lacrosse head perpendicular to the ball's velocity vector.
We capture both these properties via both hard and soft constraints.

\paragraph{Hard endpoint constraint $c$.}
The endpoint catching constraints capture the requirement that the lacrosse head must be positioned and oriented correctly to accept the incoming projectile. In particular, let $(p_o(t_f), v_o(t_f))$ be the true 3D position and velocity of the object at the catching time $t_f$. Then, we require:
\begin{equation}
    \| p_{h}(q(t_f)) - p_o(t_f) \| \leq \epsilon_p, \quad \text{and} \quad 
    (R_{h}(q(t_f)) e_2)^T \dfrac{v_o(t_f)}{\|v_o(t_f)\|} \geq \cos \epsilon_r,
\label{end_constraints}
\end{equation}
where $\epsilon_p, \epsilon_r \in \R_{>0}$ are prescribed tolerances on the position and angular errors, respectively, and $e_2 = (0, 1, 0)^T$. The second constraint enforces alignment of the local $\hat{y}-$axis on the lacrosse head, which is orthogonal to the net's catching plane, and the ball's velocity vector at $t_f$. 

The constraint above is written assuming access to the ball's true 3D position and velocity. However, since we only have access to a prediction of these quantities via the parametric predictor $\hat{F}_o(\cdot; \theta_o)$, we enforce the above constraints w.r.t. the predicted quantities $\hat{p}_o(t_f; \theta_o), \hat{v}_o(t_f; \theta_o)$, making the endpoint catching constraint function $c(q(t_f), t_f; \theta_o)$ parametric in $\theta_o$.

\paragraph{Soft terminal cost $\Psi$.}
In conjunction with the hard constraints above, the terminal cost $\Psi$ takes the following form:
\begin{align}
    \Psi(q(t_f), \dot{q}(t_f), t_f; \theta_o) &:= w_p\psi_p(q(t_f), t_f; \theta_o) + w_v\psi_v(q(t_f), \dot{q}(t_f)) \\
    \psi_p(q(t_f), t_f; \theta_o) &:= \| p_{h}(q(t_f)) - \hat{p}_o(t_f; \theta_o) \|^2 +  \left(1 - (R_{h}(q(t_f)) e_2)^T \dfrac{\hat{v}_o(t_f; \theta_o)}{\|\hat{v}_o(t_f; \theta_o)\|} \right) \\
    \psi_v(q(t_f), \dot{q}(t_f)) &:=  \left\| R_h(q(t_f))^T v_{h}(q(t_f), \dot{q}(t_f)) - \begin{bmatrix} 0 \\ v_c \\ 0 \end{bmatrix} \right\|^2, \label{eq:term_vel}
\end{align}
where $w_p, w_v \in \R_{\geq 0}$ are constant weights, and $v_{h}(q(t_f), \dot{q}(t_f)) \in \R^3$ is the lacrosse head's translational velocity expressed in the inertial frame, and computed via the Jacobian-vector product $\partial_q p_{h}(q) \dot{q}$. The constant $v_c \in \R$ is a desired catching \emph{speed}. Thus, the terminal cost $\Psi$ penalizes the catching-time pose errors, as defined within~\eqref{end_constraints}, as well as the motion of the lacrosse head perpendicular to the ball's velocity vector at the catching instant.

The overall OCP is thus parametric in $\theta_o$, the parameters of the ball's 3D predictor function $\hat{F}_o$, and problem parameters $\{\epsilon_p, \epsilon_r, v_c, w_p, w_v, \lambda\}$.

\section{Reduction to Sequential Quadratic Programming}

The OCP \eqref{ocp_obj} is a non-trivial problem which could be solved by leveraging the necessary conditions of optimality for free end-time problems and using boundary-value-problems solvers. However, this would entail optimizing over control, state, and co-state trajectories using dense discretization of the dynamics and inequality constraints (e.g., via collocation). Instead, we simplify the computational burden by optimizing over a restricted class of solutions -- a sequence of acceleration and coasting phases, and in the process, convert the problem into a \emph{multi-stage} discrete-time trajectory optimization problem that is subsequently solved using a state of the art shooting-based Sequential Quadratic Programming (SQP) solver~\citep{singh_sqp_2022}. 

\subsection{Conversion to Multi-Stage Trajectory Optimization}

We assume that the acceleration limits are given by symmetric intervals $[-\ddot{q}_a, \ddot{q}_a]$, where $\ddot{q}_a \in \R^7_{>0}$ is a fixed vector. Then, we can define an $N$-stage discrete-time trajectory optimization problem, where each ``stage" is composed of a constant acceleration phase followed by constant cruise phase. Formally, stage-$k$ for $k \in \{0,\ldots,N-1\}$ lasts for $\delta t[k]$ seconds, where $\delta t[k] \in \R_{\geq 0}$. Then, within the acceleration phase of stage-$k$, joint $i \in \{1,\ldots, 7\}$ accelerates at $\pm \ddot{q}_{a_i}$ starting at $(q_i, \dot{q}_i)[k]$ to achieve a net velocity change of $\delta \dot{q}_i[k]$. In the cruise phase, the joint moves at a constant rate of $\dot{q}_i[k] + \delta \dot{q}_i[k]$ for $\delta t[k] - (|\delta \dot{q}_i[k]| / \ddot{q}_{a_i})$ seconds. We can summarize the stage transition above by defining a composite state $x$ and control $u$:
\[
    x[k] := (q[k], \dot{q}[k], t[k]) \in \R^{15}, \quad u[k] := (\delta \dot{q}[k], \delta t[k]) \in \R^{8}
\]
Then, the stage-``dynamics" are written as:
\begin{equation}
    x[k+1] = \begin{bmatrix} q[k+1] \\ \dot{q}[k+1] \\ t[k+1] \end{bmatrix} = \begin{bmatrix} q[k] + (\dot{q}[k] + \delta \dot{q}[k]) \delta t[k] - (1/2)\Delta_{\ddot{q}_a}^{-1}\left(\delta \dot{q}[k] \circ |\delta \dot{q}[k]|\right) \\
    \dot{q}[k] + \delta \dot{q}[k] \\
    t[k] + \delta t[k] \end{bmatrix}
\label{stage_dyn}
\end{equation}
where $\circ$ denotes the Hadamard product, and $\Delta_v$ is the diagonal matrix form of the vector $v$. Let $\bm{u}:=(u[0],\ldots, u[N-1])$. The stage-equivalent discrete-time objective is given as:
\begin{equation}
    J(\bm{u}) = \sum_{k=0}^{N-1} \left(\lambda \delta t[k] + \|\delta \dot{q}[k]\|^2 \right) + \Psi(x[N]).
\label{ocp_obj_dt}
\end{equation}

\begin{remark}
Note that the exact conversion of the integral objective in~\eqref{ocp_obj} to the stage-wise discrete-time objective would result in a stage-cost of the form $\lambda \delta t[k] + \ddot{q}_a^T | \delta \dot{q}[k]|$. However, this was found to be numerically less robust than the $\mathcal{C}^2$ smooth objective used above.
\end{remark}

The terminal cost and endpoint catching inequality constraints from~\eqref{end_constraints} carry over directly, and are applied to $x[N] = (q(t_f), \dot{q}(t_f), t_f)$, where $t_f = \sum_{k=0}^{N-1} \delta t[k]$. We now tackle the limit constraints on $(q, \dot{q}, \ddot{q})$. For acceleration, we require:
\begin{equation}
|\delta \dot{q}[k] | \leq \ddot{q}_a \delta t[k], \quad k = 0,\ldots, N-1.
\end{equation}
Since $\dot{q}(t)$ linearly interpolates between the stage-values $\dot{q}[k]$, the velocity limit constraints need only be enforced at the stage values:
\begin{equation}
    \underline{\dot{q}} \leq \dot{q}[k] \leq \overline{\dot{q}}, \quad k = 0,\ldots, N.
\end{equation}
Finally, to handle the limit constraints on $q(\tau)$ for all $\tau \in [0, t_f]$, we must account for both the parabolic (constant acceleration) and linear (cruise) profiles within each stage:
\begin{itemize}
    \item {\bf Case 1}: $\dot{q}_i[k] (\dot{q}_i[k] + \delta \dot{q}_i[k]) \geq 0$. In this case $q_i(\tau)$ interpolates in-between $\{q_i[k], q_i[k+1]\}$ for all $\tau \in [t[k], t[k+1]]$. Thus, we need only apply the limit constraints on the endpoints $q_i[k], q_i[k+1]$.
    \item {\bf Case 2}: $\dot{q}_i[k] (\dot{q}_i[k] + \delta \dot{q}_i[k]) < 0$. In this case, there is a local max/min for $q_i(\tau)$ within $[t[k], t[k+1]]$ where $\dot{q}_i(\tau) = 0$. Denote this max/min as $\hat{q}_i[k]$. Then, in addition to enforcing the limit constraints at $q_i[k], q_i[k+1]$, we must also enforce the constraint on $\hat{q}_i[k]$. The expression for $\hat{q}_i[k]$ is given by:
    \[
        \hat{q}_i[k] = q_i[k] + \begin{cases} \frac{\dot{q}_i[k]^2}{2 \ddot{q}_{a_i}} \quad &\text{ if } \dot{q}_i[k] > 0 \\
        -\frac{\dot{q}_i[k]^2}{2 \ddot{q}_{a_i}} \quad &\text{ if } \dot{q}_i[k] < 0
        \end{cases}.
    \]
\end{itemize}

Given the discrete-time ``stage"-dynamics, optimization objective, and constraints, we can use any off-the-shelf constrained discrete-time trajectory optimization solver. In this work, we leverage Dynamic Shooting SQP, introduced in~\cite{singh_sqp_2022}.

\begin{remark}
Note that the combination of max/min acceleration and cruise phases within each stage reflects the nature of mixed control-effort/minimum-time optimal control solutions, colloquially characterized as the ``bang-off-bang'' strategy. Recent work~\citep{sarkar2021characterization} has shown that for LTI systems with a single control input, the optimal solution to a mixed control-effort/minimum-time problem with an endpoint reachability constraint is a sequence of ``bang-off'' stages. This justifies our use of such a stage-wise reduction of the original continuous-time OCP, and is similar in spirit to previous works on catching using trapezoidal velocity profiles~\citep{bauml2010kinematically}.
\end{remark}

\subsection{Further implementation details}

We now describe further implementation details
of our SQP implementation.

\paragraph{Asynchronous Implementation:} Running concurrently to the catching controller is an estimator that generates updates of the predictor parameters $\theta_o$, necessitating online re-planning. We achieve this via an asynchronous implementation where the optimization problem is continually re-solved in a separate thread, using the latest estimate for $\theta_o$ and the current robot state $(q, \dot{q})$. The commanded $(q, \dot{q})$ for the robot's lower-level PD controllers are computed by decoding the most recent stage-wise solution to a continuous-time trajectory, thereby guaranteeing a consistent control rate. We note that since there is no receding horizon, re-planning is more akin to fine-tuning of the plan in response to the updating estimate of the ball's trajectory, as opposed to traditional model-predictive-control. Thus, in the absence of errors in the ball's trajectory's prediction, the problem remains recursively feasible.

\paragraph{Cradling:} Following the intercept of the ball, we use an \emph{open-loop} cradling motion primitive, modeled as a 2nd-order ODE in $q$, to slow the lacrosse head and simultaneously rotate the net to point upwards. In particular, for $t \geq t_f$, we let $\ddot{q}(t) := u_c(t, q(t), \dot{q}(t))$, where $u_c$ is an acceleration controller, detailed below. Define $\hat{y}_h(q) := R_h(q)e_2$, i.e., the end-effector frame's $\hat{y}-$axis. Then, for $t \geq t_f$, let $\nu(t) := \min \{ (t - t_f)/t_s, 1\}$, where $t_s \in \R_{>0}$ is a user-set constant, denoted as the ``slow-down" time. Then, we define the \emph{desired} translational $v_d$ and rotational $\omega_d$ velocity for the lacrosse head as follows:
\begin{align*}
    v_d(q(t), t) &:= v_c (1 - \nu(t)) \cos(\pi \nu(t)) \hat{y}_h(q(t)) \\
    \omega_d(q(t)) &:= -\pi (\hat{y}_h(q(t)) \times e_3),
\end{align*}
where $e_3 := (0, 0, 1)^T$, and $v_c \in \R$ is the desired catching speed defined in~\eqref{eq:term_vel}. Thus, the desired translational velocity $v_d$ is aligned with the end-effector's $\hat{y}-$axis, and slows the head down from $v_c$ to $0$. Meanwhile, the desired rotational velocity tries to align the net to point upwards. The combination of these velocities results in a ``scooping-like" motion, intended to cradle the ball. The desired acceleration is then given by simple proportional feedback:
\begin{equation}
    \dot{v}_h = k_v (v_d - v_h(q, \dot{q})), \quad \dot{\omega}_h = k_{\omega}(\omega_d - \omega_h(q, \dot{q})),
\label{vel_lpf}
\end{equation}
where $\omega_h(q, \dot{q})$ is the lacrosse head's rotational velocity expressed in the inertial frame, and $k_v, k_{\omega} \in \R_{>0}$ are user-set constant gains. To convert these accelerations into the joint accelerations $u_c$ at each sampled control step, we integrate~\eqref{vel_lpf} from the current $(v_h(q, \dot{q}), \omega_h(q, \dot{q}))$ using Euler integration for one controller time-step, and invert the Jacobian of the FK transform at $q$ to compute an updated set of desired joint velocities $\dot{q}^+$. The resulting desired change in velocity, $\dot{q}^+ - \dot{q}$ is then clipped according to the acceleration constraints, yielding the final joint acceleration $u_c$. Finally, we perform one controller time-step integration of $\ddot{q} = u_c$ assuming a zero-order-hold on $u_c$, to update $(q, \dot{q})$. 

\section{Blackbox Gradient Sensing Optimization}

The catching problem \eqref{ocp_obj} 
can also be formulated as a 
Partially-Observable Markov Decision Process (POMDP), defined by the tuple $(S, O, A, R, P)$ where $S$ is the state space partially observed by $O$, the observation space, $A$ is the action space, $R: S \times A \mapsto \mathbb{R}$ is the reward function and $P: S \times A \mapsto S$ is the dynamics function. The optimization objective is to learn a parameterized policy $\pi_{\theta}: O \mapsto A$ that maximizes the expected total episode return, $J(\theta) = \mathbb{E}_{\tau=(\mathbf{s}_0, \mathbf{a}_0, \dots, \mathbf{s}_T)} \sum_{t=0}^{T} r(\mathbf{s}_t, \pi_{\theta}(\mathbf{o}_t))$. The state space consists of the ball position, velocity, and a predicted trajectory, which are approximated from the raw observations generated by the perception system and an imperfect dynamics model, thereby justifying the POMDP categorization. In this work, we optimize a neural network policy via Blackbox policy optimization \citep{ICML-2018-ChoromanskiRSTW, rbo, MGR2018}.

\subsection{Reward design}

Careful reward design is necessary to ensure the success
of blackbox policy optimization. 
The reward function is different for training in sim vs.\ real due to differences in quality of data from each. In both cases we reward the net getting close to the ball during the episode. In sim, we additionally reward orientation alignment before the catch + a stability reward for keeping the ball in the net; in real, we use a flat reward for successful catches (detected by a sensor). Finally, we discourage excessive motion via penalizing position/velocity/acceleration/jerk in sim, and hardware limit violation in real. All terms are summed together to yield the net reward. In more detail:
\begin{itemize}
    \item  \textbf{Object Position Reward} (sim/real): 
    We compute the closest distance the end-effector comes to the object during the episode. The closest distance is scaled on an exponential curve with a cutoff at $20cm$, scoring $1.0$ for any episodes below this threshold. 

    \item \textbf{Object Orientation Reward} (sim): 
    We compute the orientation of the net right when the ball is within $20cm$ of the net. The score is computed as a dot product of the velocity vector of the object and the axis of the net, scaled between $0$ to $1$.

    \item \textbf{Object Stability Reward} (sim): 
    This reward computes how stable the object remains after it is close to the net (within $20cm$ of the net). Entering the close criteria and staying there through the end of the episode provides a flat $0.2$ reward. The remaining $0.8$ part of the stability reward is given by measuring the speed of the ball while it's close for $0.25s$. Each time-step during this duration contributes equally and is scored on an exponential curve based on object speed, capping out at speeds less than $0.2m/s$ scoring full for that timestep. A full score keeps the speed less than $0.2m/s$ for the full $0.25s$. This reward is only used in sim as the precision of ball tracking in real
    is difficult when the ball is in the net or obscured.

    \item \textbf{Object Catch Reward} (real): 
    We use a proximity sensor attached close to the net that can reliably detect whether a ball is in the net or not. The ball is declared as caught if the sensor detects a ball continuously in the net for greater than $0.25s$. This provides a flat $0$ or $1$ reward.

    \item \textbf{Penalties for exceeding dynamic constraints} (sim):
    We use multiple penalty rewards used to ensure the policy learns to operate within the robot constraints such as joint position, velocity, acceleration and jerk limits. The penalty rewards are implemented as a flat $1.0$ if the agent actions stay within constraints and reduces to $0$ depending on how much it violates them. The reward is reduced depending on how many timesteps and by how much it exceeds them. 

    \item \textbf{Penalties for exceeding dynamic constraints} (real):
    The hardware produces a fault error code and freezes when movements exceed constraints. Thus, we assign
    a binary penalty whether the hardware encounters the fault code or not.
    
\end{itemize}

While the \emph{Object Catch Reward} is a direct measure of catch success, it is a sparse metric for optimization, particularly during early stages of training. Consequently, the \emph{Object Position} and \emph{Object Orientation} reward terms provide more dense guidance, encouraging the policy to align the net correctly with the catching pose and start learning. The \emph{Object Stability} reward was necessary to penalize interactions where the ball simply bounced off the net. Finally, the \emph{Dynamic constraint penalties} were necessary for successful transfer to the real robot in order to respect physical constraints.

\subsection{Further implementation details}

We now describe further implementation details of our blackbox policy optimization.

\paragraph{Policy Network:} We use a two-tower CNN neural network. The first tower process the historical joint positions represented as an image of size $(n_{\mathrm{hist}}, 7)$, where $n_{\mathrm{hist}}$ is the number of past timesteps. The second CNN tower process the predicted ball trajectory represented as an image of size $(n_{\mathrm{pred}}, 6)$, where $n_{\mathrm{pred}}$ is the number of predicted timesteps. The output of the two towers is concatenated into a single tensor, which is fed into two fully-connected layers. %
The final output is then taken as the commanded joint velocities.
In total, our policy network has $3255$ parameters.

\paragraph{Blackbox Gradient Sensing and Sim-to-Real Finetuning:} We apply
Blackbox Gradient Sensing (BGS) for optimizing the policy neural network parameters $\theta$, due in part to both the simplicity of the method and recent successes in a variety of robotic domains \citep{abeyruwan2022sim2real}. The algorithm optimizes a smoothened version $J_{\sigma}(\theta)$ of the original total-reward objective $J(\theta)$, given as: $J_{\sigma}(\theta) = \mathbb{E}_{\mathbf{\delta} \sim \mathcal{N}(0,\mathbf{I}_{d})}[J(\theta+\sigma \mathbf{\delta})]$,
where $\sigma > 0$ controls the precision of the smoothing, and $\delta$ is an
isotropic random Gaussian vector.
We first train in a simulation environment implemented in 
PyBullet~\citep{coumans2021}.
Once the policy performs well in simulation, we transfer the policy to the
real robot and run further BGS finetuning steps using the mechanical thrower.

\section{Experiments}
\label{sec:experiments}

We evaluate both our SQP and blackbox (BB) agents in simulation, on the real robot, and also explore performance
under various distribution shifts of the thrower.
Our SQP agent uses a state of the art SQP solver~\citep{singh_sqp_2022} built on top of trajax~\citep{trajax2021github}, a JAX library for differentiable optimal control. For the SQP agent, we found that a single stage, i.e., $N=1$ in~\eqref{ocp_obj_dt}, was sufficient to achieve a high catching success rate, albeit with less flexibility on matching the desired catching speed defined in the soft terminal cost in~\eqref{eq:term_vel}. For our BB agent, we use a distributed BGS library~\citep{choromanski2018structured} with policy networks implemented in Tensorflow Keras. The robot used is a combination of an ABB IRB 120T 6-DOF arm mounted on a one-dimensional Festo linear actuator, creating a 7-DOF system. The ball location is determined using a stereo pair of Ximea MQ013CG-ON cameras running with a trained recurrent tracker model. 

\paragraph{Error bars:}
We show 
catch success for the real robot with error bars which
give at least $95\%$ coverage, by using 
the Clopper–Pearson method to compute binomial 
confidence intervals.

\paragraph{Inference speed and Reaction time:}
The BB agent 
computes a single policy action in time $7.253$ ms (std.~$0.160$ ms),
whereas SQP takes $43.046$ ms to solve (std.~$21.255$ ms). Recall that the SQP runs asynchronously,
so this solve time does not block the agent; the synchronous part 
runs in $2.139$ ms (std.~$0.212$ ms). Vision/hardware joint data processing takes about $5$ ms. Overall agents are set to synchronously run at 75Hz. The mechanical thrower is $3.9$ meters away from the robot and imparts $4.5$ m/s horizontal velocity alone; including the $z$-component the speed is $\sim 5.5$ m/s at catch time.

\begin{figure}[!htb]
\begin{minipage}{0.32\textwidth}
  \includegraphics[width=\linewidth]{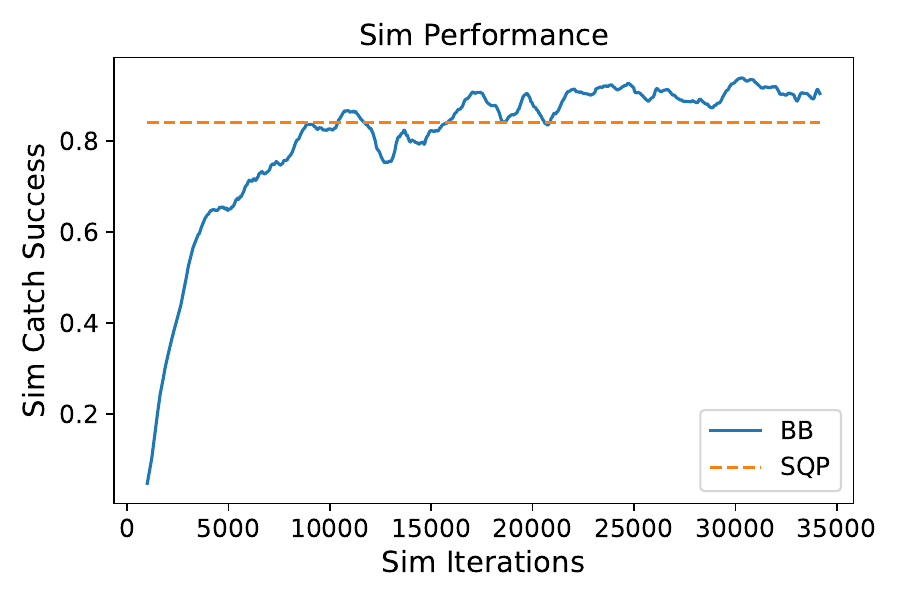}
\end{minipage}\hfill
\begin{minipage}{0.32\textwidth}
  \includegraphics[width=\linewidth]{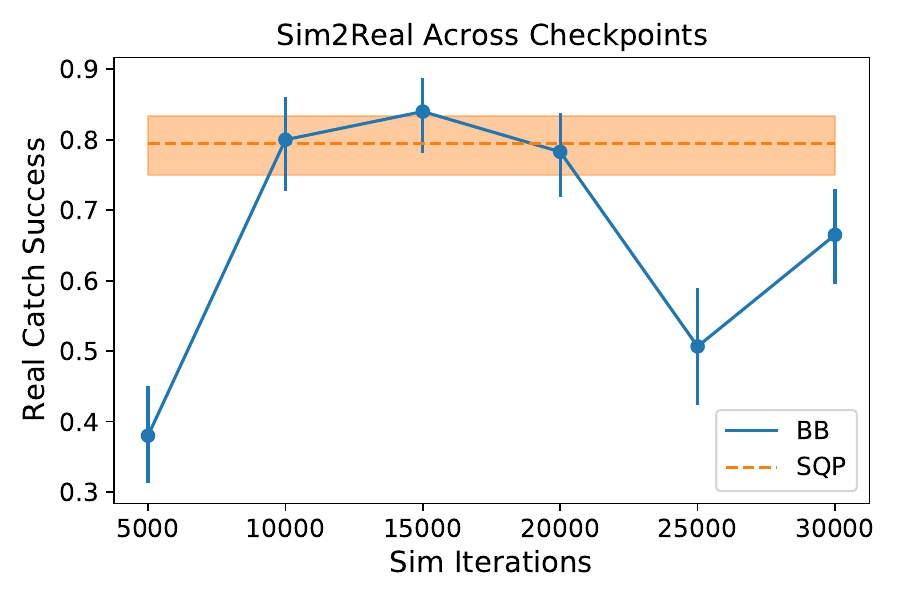}
\end{minipage}\hfill
\begin{minipage}{0.32\textwidth}%
  \includegraphics[width=\linewidth]{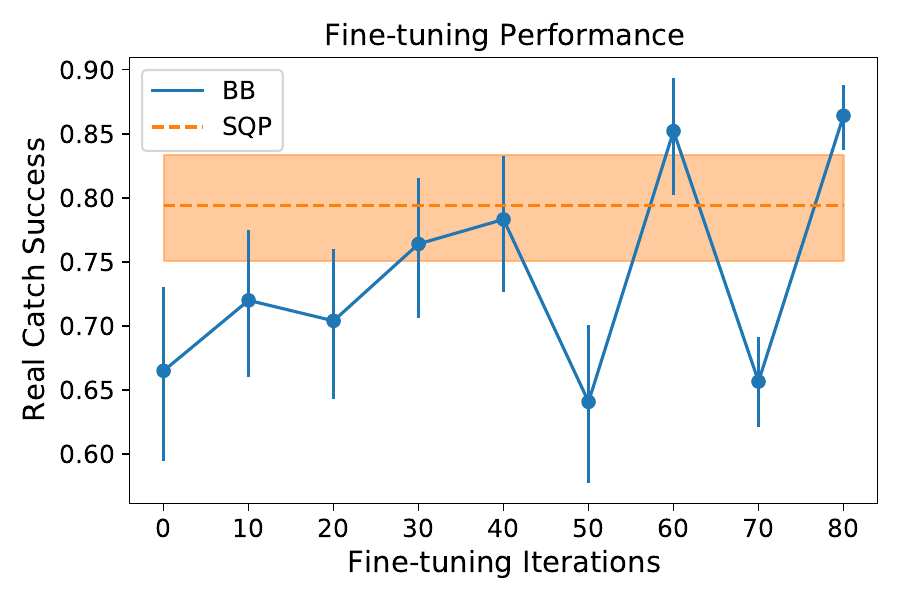}
\end{minipage}
\caption{
\textbf{(Left)} Performance of agents in sim.
\textbf{(Middle)} Performance of agents on real without fine-tuning.
\textbf{(Right)} Performance of sim2real transfer after fine-tuning the BB agent starting from the 30k iteration sim checkpoint.
Note that each iteration corresponds to 30 mechanical ball throws. The higher variance of the BB fine-tuned policy is a consequence of using a significantly smaller number of throws per BGS gradient step on real (30) as compared to simulation (100).
}
\label{fig:sim2real}
\end{figure}

\paragraph{Simulation to Reality Transfer:} \Cref{fig:sim2real} highlights the 
real robot catch performance of both
SQP and BB agents.
First, we see that while BB performance in 
sim is mostly monotonically increasing (\Cref{fig:sim2real}, left), this does not
necessary translate to monotonic improvement on the real hardware (\Cref{fig:sim2real}, middle).
Secondly, we see that SQP suffers less performance degradation
compared to BB when transferring to real.
Finally, we see that it takes 40 iterations of fine-tuning
on real (30 ball throws per iteration) in order for the fine-tuned BB
agent to match SQP's real performance (and eventually
exceed it). Both methods achieve about 80 to 85\% success on mechanical ball throws.

\begin{figure}[!htb]
\begin{minipage}{0.32\textwidth}
  \includegraphics[width=\linewidth]{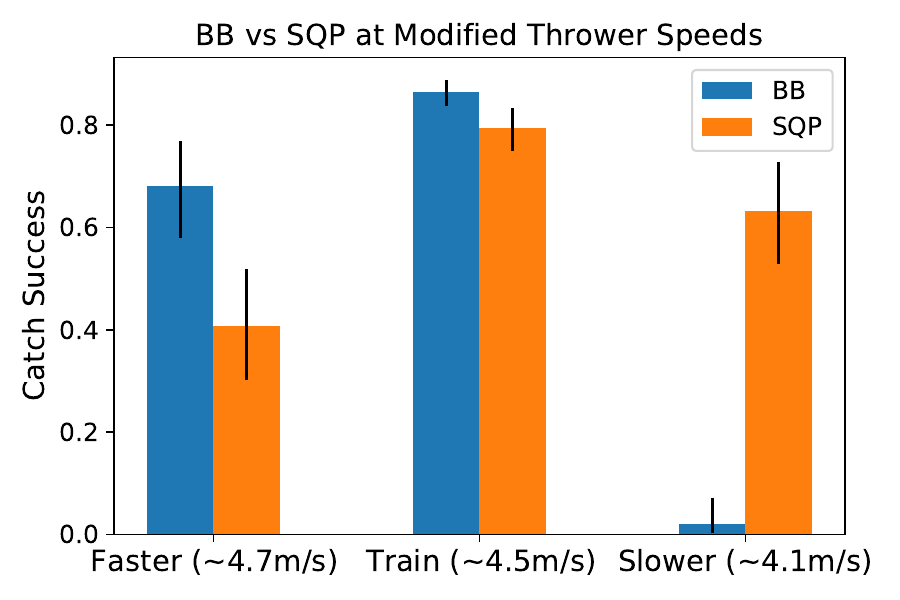}
\end{minipage}\hfill
\begin{minipage}{0.32\textwidth}
  \includegraphics[width=\linewidth]{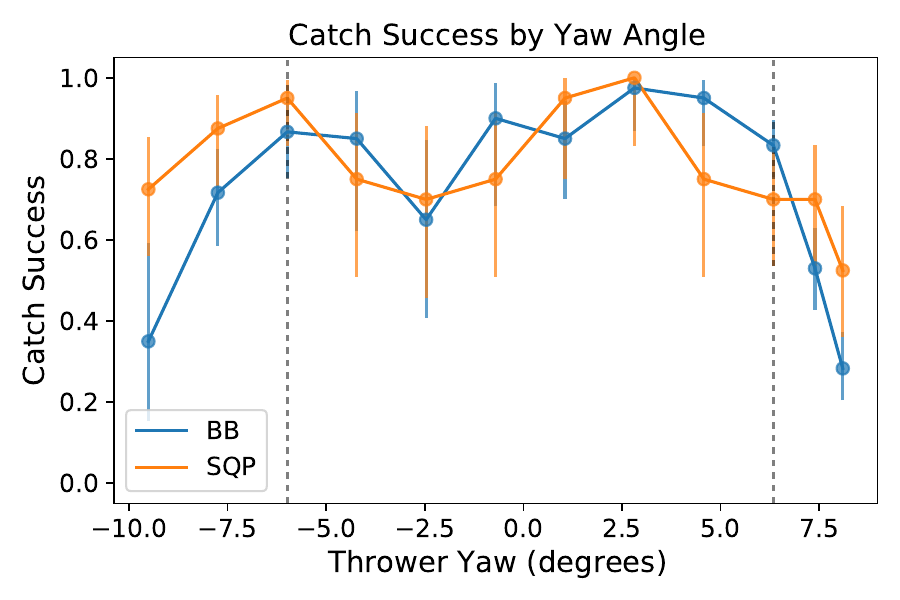}
\end{minipage}\hfill 
\begin{minipage}{0.32\textwidth}
  \includegraphics[width=\linewidth]{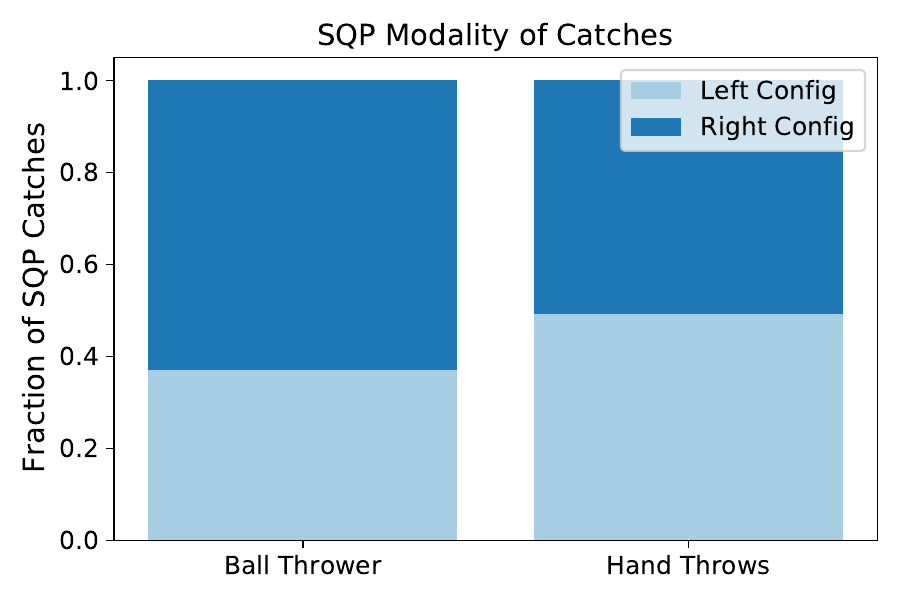}
\end{minipage}
\caption{\textbf{(Left)} Catch performance as thrower speed varies
between faster ($\sim 4.7$ m/s), training ($\sim 4.5$ m/s), and slower ($\sim 4.1$ m/s) throws.
\textbf{(Middle)} Catch performance as the thrower yaw angle varies
from $-9.5^{\circ}$ to $8^{\circ}$. Note that the training distribution
varies between $-6^{\circ}$ and $6.3^{\circ}$ (marked by the dashed vertical black line).
\textbf{(Right)} Distribution of left and right catches by the SQP agent
on both mechanical ball throws and hand throws. Note that the
BB agent catches to the right 100\% of the time,
likely due to the learning bias from the ball throw distribution.
}
\label{fig:ood}
\end{figure}

\paragraph{Robustness to Distribution Shifts:} Next, we look at the robustness of both agents to
out-of-distribution throws. For BB, this is the post-fine-tuned on real policy. We consider three different
distribution shifts: (i) varying the speed of the thrower,
(ii) varying the yaw angle of the thrower, and
(iii) throwing balls by hand instead of using the mechanical thrower.
The first two distribution shifts are plotted in
\Cref{fig:ood}. In \Cref{fig:ood} (left), we see that while
BB is reasonably robust to faster throws, its performance significantly degrades for slower throws. This is in contrast to the SQP
agent, which moderately degrades in performance for faster throws (most likely
due to computational bottlenecks), but is quite robust to slower throws.
In \Cref{fig:ood} (middle), we see that both agents have similar performance across the in-distribution yaw angles, but for out-of-distribution angles SQP maintains its performance better relative to BB.

Our last distribution shift involves hand throws (lobs) to the thrower instead of using the mechanical thrower. 
Using hand throws, the SQP agent has a 68.9\% catch success (over 196 throws), whereas the BB agent catch performance degrades to 2.0\% (over 150 throws). While the BB policy can be further fine-tuned on the hand-thrower distribution, the number of throws required would be prohibitively time-consuming.

\paragraph{Multimodality:}
In \Cref{fig:ood} (right), we demonstrate that the SQP agent
is able to catch balls in both a left and right pose configuration
at fairly even rates matching the bias of the thrower.
On the other hand, the BB agent is only able to catch to the right, since the ball thrower distribution is biased (60/40\%) towards
throwing to the right. We intuit that during the early phase of training the policy exploits the split in throw distribution to learn a right hand side catching behavior which is locally optimal and later fine-tunes this strategy to catch the left hand side balls by moving the base.

\section{Conclusion and future work}
\label{sec:conlusion}
While the fine-tuned blackbox agent has the highest catching
success performance, the SQP agent is much more robust to distribution shifts
in the thrower. To obtain the ``best of both'', we plan to investigate
the following strategies combining blackbox optimization and SQP:
\begin{itemize}
    \item Use blackbox optimization (via CMA, BGS, etc.)
    to optimize the set of tuneable SQP and cradling parameters $\theta$.
    \item Optimize a (smaller) BB policy to output SQP and cradling parameters $\theta$ for each episode, conditioned on ball history and current proprioception.
    \item Study a ``switch-over'' policy, where 
    SQP is followed until a switch-over point where control is handed over
    to the BB policy right before intercept and cradling. The BB policy additionally outputs a binary variable indicating when to switch. This limits the complexity of the BB policy to just capturing the correct ``cradling'' behavior.
\end{itemize}
Future extensions also include introducing tools from adaptive nonlinear dynamics prediction, for applications such as catching of light balls with significant aerodynamics (e.g., quadratic drag and Magnus effects), as well as catching of larger non-spherical objects.

{

}

\newpage
\appendix

\section{Author contributions}

\textbf{Saminda Abeyruwan, Alex Bewley, David D'Ambrosio:} 
Implemented the vision system and Kalman filtering.
\\
\textbf{Krzysztof Choromanski:} Designed (with Deepali) 
the Blackbox Gradient Sensing algorithm applied in all Blackbox training runs. Wrote the Blackbox optimization section of the paper.
\\
\textbf{Deepali Jain:} Designed the two-tower CNN policy architecture for the BB agent; Integrated BB policy with adaptive predictor for ball trajectory observations and ran sim experiments to reach 90\% catch success.
\\
\textbf{Anish Shankar:} Designing \& running experiments, analyzing results for insights into hardware, environment, agent improvements; Collaborating with the rest of the team on iterating the research loop for better agent directions; Designed \& Developed the catching environment along with hardware integration \& designing suitable agent observations/rewards.
\\
\textbf{Sumeet Singh:} Designed, debugged, and polished (in sim) the optimal control formulation and SQP reduction; Wrote the code for the synchronous SQP agent, and paired (with Stephen Tu) for the asynchronous adaptation; Iterated upon real experiments with Anish; Wrote the problem formulation, optimal control, SQP theory, and cradling sections of the paper, along with overall editing.
\\
\textbf{Pannag Sanketi:} Co-founded and managed the project. Advised on the research direction, experiments and the paper story. Wrote parts of the paper.
\\
\textbf{Vikas Sindhwani:} Wrote the differentiable kinematics and iLQR routines used within the SQP solver; Contributed to the implementation of simulation environment; Conducted early BlackBox experiments; Contributed to paper writing and provided overall project guidance and direction.
\\
\textbf{Jean-Jacques Slotine:} Advised on research direction and provided project guidance.
\\
\textbf{Stephen Tu:} Wrote asynchronous catching SQP implementation, and debugged performance issues in the SQP agent to make it practical for real time use; Wrote the experimental section of paper along with Anish.

\end{document}